\documentclass[runningheads,a4paper]{llncs}

\usepackage{amssymb}
\usepackage{url}

\usepackage{graphicx,times,epsfig,amsmath}
\usepackage{ctable}
\usepackage{rotating}
\usepackage{multirow}
\usepackage{listings}
\usepackage{algorithm}
\usepackage{algorithmic}
\usepackage{authblk}

\urldef{\mailsa}\path|{Gopinath.Chennupati}@ul.ie|  
\urldef{\mailsb}\path|{f.e.b.otero}@kent.ac.uk| 
\urldef{\mailsc}\path|{xxxx, xxxx, xxxx}@xxxxx|  
\newcommand{\keywords}[1]{\par\addvspace\baselineskip
\noindent\keywordname\enspace\ignorespaces#1}

\begin{document}

\mainmatter  

\title{eAnt-Miner : An Ensemble Ant-Miner to Improve the ACO Classification}

\titlerunning{eAnt-Miner}


\author{Gopinath Chennupati}
\affil{Computer Science and Information Systems Department \\
	 University of Limerick, Limerick \\
	 Ireland. \\ 
\mailsa
}

\authorrunning{Ensemble Ant-Miner to Improve the ACO Classification}

\institute{}


\toctitle{eAnt-Miner}
\tocauthor{Ensemble of Ants}

\maketitle

\begin{abstract}
Ant Colony Optimization (ACO) has been applied in supervised learning in order to induce classification rules as well as decision trees, named Ant-Miners. Although these are competitive classifiers, the stability of these classifiers is an important concern that owes to their stochastic nature. In this paper, to address this issue, an acclaimed machine learning technique named, ensemble of classifiers is applied, where an ACO classifier is used as a base classifier to prepare the ensemble. The main trade-off is, the predictions in the new approach are determined by discovering a group of models as opposed to the single model classification. In essence, we prepare multiple models from the randomly replaced samples of training data from which, a unique model is prepared by aggregating the models to test the unseen data points. The main objective of this new approach is to increase the stability of the Ant-Miner results there by improving the performance of ACO classification. We found that the ensemble Ant-Miners significantly improved the stability by reducing the classification error on unseen data. 

\end{abstract}

\section{Introduction}
Data mining, the process of automatically extracting useful information from the real-world data bases. It has tremendous success in solving knowledge extraction problems from industry, science, engineering to government. Knowledge extraction can be done in many ways: clustering, classification, and regression, etc., of which classification has got the highest contribution and we focus on it in this paper. 

In essence, classification discovers a model that represents a predictive relationship between the values and the attributes in deciding the classes of a given training data set, in turn, it is used to classify the unseen data (test set). The main aim of the classification task is to improve the accuracy and building the most reliable classifier. Predictive accuracy is the most commonly used evaluation measure. 

Ant Colony Optimization (ACO)~\cite{Dorigo:aco} - a meta heuristic designed to solve the combinatorial optimization problems that has been inspired from the foraging and pheromone trails behaviour of the natural ants. ACO has been applied for the classification task, in which, Ant-Miner proposed in~\cite{Parpinelli:acomining} is the first of its kind of rule induction algorithm that deals with the nominal attributes of a given data set. Following the inception of the Ant-Miner, several extensions have been proposed as surveyed in~\cite{Martens:survey}. A potential concern with ACO classification is the stability of its results. A classifier is said to be stable if it's ability of prediction is invulnerable even for small changes in the training data. For example, for $99$ data points a stable classifier discovers a similar model as that on $100$ data points. 

Some single classifiers such as decision trees~\cite{Quinlan:c45} produce most unstable models. Quite often the accuracy and/or the stability of the single classifier has been improved with the help of ensembles. Ensemble is the process of gathering the opinion from multiple experts. Some of the acclaimed ensemble approaches are Bagging, Boosting, Stacking, Random Forests that are explained in~\cite{Witten:datamining}. For example, bagging can be described as follows: a patient consults different doctors to get the prescription from each of them and at the end, will go with the majority of the prescribed opinions through voting. This simple technique highly reduces the variance present in the predictions. 

In this paper, we propose to apply ensemble learning on Ant-Miners, named \textit{eAnt-Miner} that improves the accuracy of the ACO classification while stabilizing the classification. The basic idea behind the creation of an ensemble Ant-Miner is to create a bagged classifier. We examine the stability of different Ant-Miners in the literature before we evaluate the ensemble approach. Then, we create a bagged Ant-Miner for each of the base Ant-Miner variation. We evaluate the impact of the ensembles in comparison with the single classifiers on $17$ publicly available data sets that are drawn from the UCI Machine Learning repository~\cite{Bache:uci}. The evaluation is carried both in terms of the classification error and the size of the discovered model.

The organization of the rest of the paper is as follows. We describe different search strategies of the Ant-Miner variations in section~\ref{relatedwork}. Next, we examine the stability of Ant-Miners in section~\ref{stability} and, in section~\ref{eantminer}, we explain the proposed ensemble approach. The experimental approach and results are explained in section~\ref{experiments}. Finally, we conclude and provide some recommendations for future work in section~\ref{conclusion}.
 
\section{Background}
\label{relatedwork}
ACO classification follows a separate-and-conquer strategy in order to induce the rules. This procedure operates iteratively in two major steps: in the first step, the algorithm creates rules that classify a part of the training set (conquer), whereas in the second step, it removes the covered data points from the data set (separate). The inductive rule based classification models are of the form \textit{IF $<term_{1}$ AND ... AND $term_{n}>$} \textit{THEN $<class>$ }, where the \textit{IF} part is the antecedent and, the \textit{THEN} part is the consequent.

In the ACO classification, a construction graph is prepared with the (attribute, value) pair taken from the data set. Then, each ant stochastically chooses a vertex based on the pheromone and the heuristic information. The chosen vertex is a rule term where the ant continues to add new terms until all the attributes are covered. This newly created rule is pruned to remove the insignificant terms. This way a rule is prepared by every ant in the colony. A best rule is selected with the help of a quality function. The pheromone levels are adjusted by increasing it on the terms included in the best rule and decreased on the remaining. The discovered rules are pruned and then, the data points that the rule covers are removed and the next rule is created. This way, the algorithm continues till the predefined number of data points are reached and at the end the best list of rules are returned as the classification model. Many of the Ant-Miner variations follow this procedure except the new ACO based algorithms proposed in~\cite{Medland:campb,fernando:ucampb}

Medland et al.~\cite{Medland:campb} proposed a new Ant-Miner, named $cAnt-Miner_{PB}$, where each ant creates a best list of rules as opposed to the single rule creation by the \textit{cAnt-Miner} in~\cite{fernando:cantminer}. This small change has dramatically improved the accuracy of the classification. In~\cite{fernando:ucampb}, an improved method named as Unordered-$cAnt-Miner_{PB}$ was introduced to discover the unordered rules (set of rules) as opposed to the ordered rules discovered by $cAnt-Miner_{PB}$. It differs slightly from $cAnt-Miner_{PB}$ search strategy by constructing a set of rules for prediction instead of creating a list of rules. In this, the set of ant discovered rules are employed on the instance and at the end a single rule is used to predict the class of a given instance, as a result it has improved the interpretability of individual rules by reducing the size of the discovered model.

We discussed rule based classification models so far, ACO can also discover decision trees. Otero et al.~\cite{fernando:trees} proposed a new decision tree induction algorithm, Ant-Tree-Miner by combining the traditional decision tree methods and ACO. It has induced the trees that are of the form similar to the trees produced by C4.5~\cite{Quinlan:c45}. Although the ACO based classifiers are proved to be competitive enough with the well established machine learning methods, they still exhibit unstable behaviour. We examine the stability of different Ant-Miners in the following section.

\section{Stability of Ant-Miners}
\label{stability}
A common notion in the machine learning community is that no classification algorithm is perfect. It can be true in many cases, in fact, it is worth interesting to test this fact, in the case of Ant-Miners. In fact, stability of the classifier suffers with a small change in the data. As all the Ant-Miners devise either the rules or the trees through the construction graphs in multiple iterations by a group of ants, operating on a different set of records in a dataset: particularly, this type of stochastic design, as a result, the inherent imperfection exerted, contributes to the unstable behaviour of Ant-Miners. With the effect from these reasons, it is important to examine the stability of Ant-Miners. 

In this section we examine the stability of four relatively new Ant-Miners (\textit{cAntMiner, $cAntMiner_{PB}$, $Unordered-cAntMiner_{PB}$, Ant-Tree-Miner}) with an observation through the \textit{test error} of each classifier on a given dataset. The variation in the test error is observed in each run of the $10$-fold cross validation~\cite{Witten:datamining} procedure. In essence, in this procedure the dataset is divided into $10$ approximately equal chunks that in which, one chunk acts as test set while the remaining $9$ chunks act as training set. It is continued for $10$ runs varying the test set in each run. In this study, we selected six benchmark data sets\footnote{Note that there is no selection criterion on the datasets which are picked randomly for these preliminary experiments.} (\textit{balance-scale, credit-g, heart-c, breast-tissue, glass, pima}) that are publicly available at UCI Machine Learning repository. 
  
\begin{figure}[htp]
  	\centerline{\includegraphics[width=12.9cm]{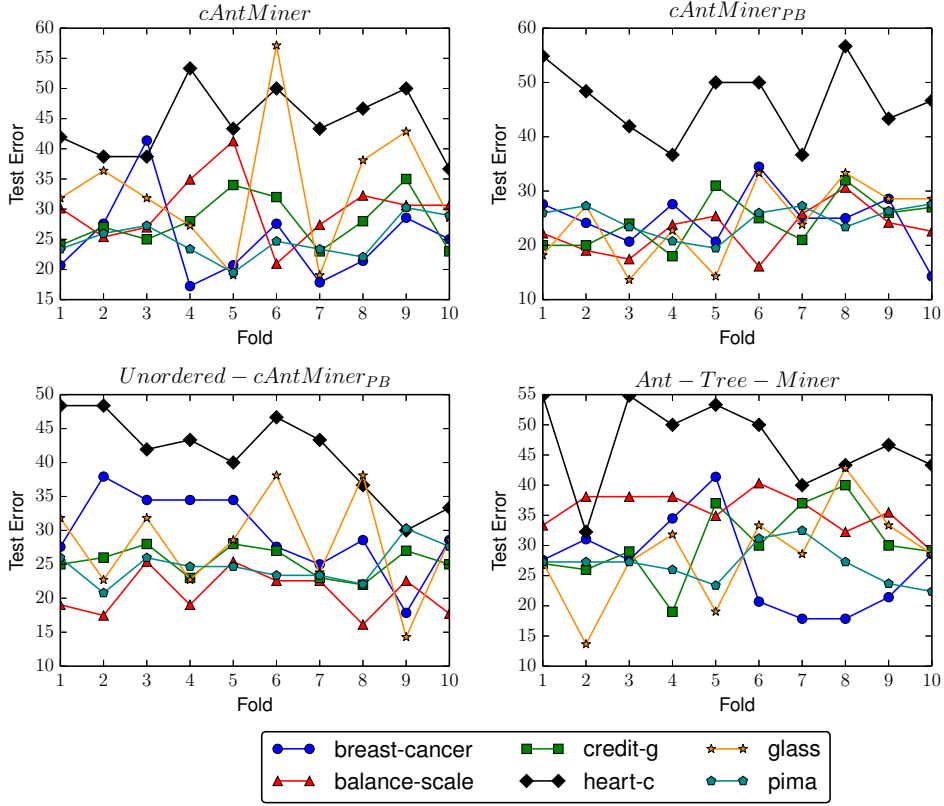}}
  	\caption{An analysis through the stability of the four Ant-Miners (\textit{cAntMiner, $cAntMiner_{PB}$, $Unordered-cAntMiner_{PB}$, Ant-Tree-Miner}) when applied $10$-fold cross validation across randomly selected datasets from the UCI Machine Learning repository. The curves represent the test error (y-axis) of the respective Ant-Miner over a given dataset with the $10$-fold cross validation.} 
  	\label{fig:stability}
  \end{figure}

Figure~\ref{fig:stability} represents the test error of the Ant-Miners in a $10$-fold cross validation (stratified cross-validation as explained in~\cite{Witten:datamining}). It shows a drastic change in the error from one run to the other on almost all the datasets. That clearly shows the unstable behaviour of all the four Ant-Miners. From the deviation in the curves, \textit{$cAntMiner_{PB}$} shows relatively less unstable behaviour while the remaining $3$ Ant-Miners are highly unstable. Considering \textit{cAntMiner}, continuous attributes are added to the construction graph that are of the form ($attribute_{i} = value_{i}$). Then, cAntMiner assimilates dynamic entropy based discretisation procedure in devising antecedents of the rule with in the boundaries of ($attribute_{i} < value_{i}$ and $attribute_{i} \geq value_{i}$). 

In the case of a small change in the dataset, a change in the boundary values is inevitable. That in fact, evolves a completely new rule with different boundary values as well as possibly a different attribute that in turn affects the predictive accuracy. For example, \textbf{if age $<$ 18 then yes}, is a rule in the first run of the $10$-fold cross validation in the next run with a small change in the data, the ants can devise a rule \textbf{if age $<$ 18 and tumour-size $\geq$ 25 then no}, that results in an unstable model with a lot of change in the predictive accuracy. Hence, it serves as a common reason behind the instability of all the dynamic entropy based Ant-Miners. Apart from this, there are other factors that can contribute to the instability as the other Ant-Miners are extended from cAntMiner, as discussed in section~\ref{relatedwork}. 

 \begin{figure}[htp]
 	\centerline{\includegraphics[width=12.25cm]{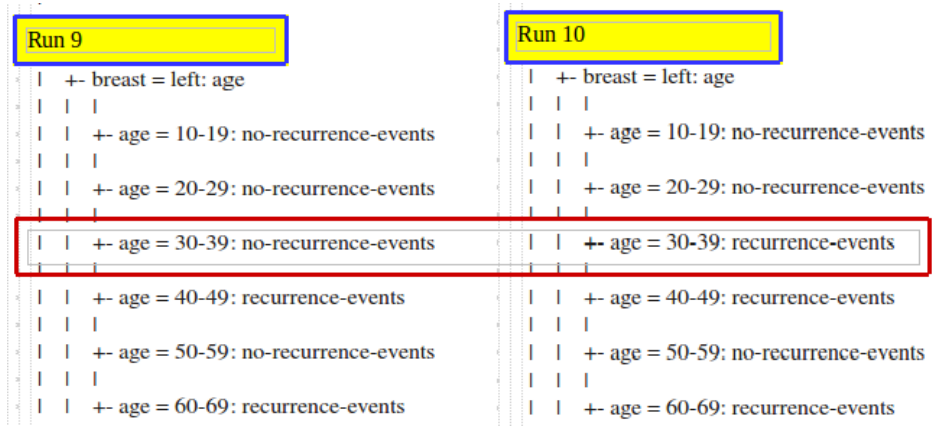}}
 	\caption{Ant-Tree-Miner discovered decision trees on the \textit{breast-cancer} dataset. The presented trees are considered from \textit{Run 9} (left) and \textit{Run 10} (right) of the same $10$-fold cross validation. Both the trees represent a portion of the whole ant devised decision trees. We focused only on the part of the tree that recorded a change in one of it's nodes.} 
 	\label{fig:stabtree}
 \end{figure} 

We now examine the remaining three Ant-Miners, of which, \textit{Ant-Tree-Miner} is highly unstable. It is because of the fact that, a small change in the condition at any of the node of the ant devised tree model expends a significant change in the accuracy of the prediction. For example, Figure~\ref{fig:stabtree} partly shows the ant devised decision trees. The highlighted rectangular box (in red) is the only change in the entire model that altered the test error (can be observed numerically in Figure~\ref{fig:stability}) to increase. This unstable behaviour is inevitable in almost all the other datasets, thus making the Ant-Tree-Miner highly vulnerable. 

In the case of $Unordered-cAntMiner_{PB}$, where the covered data points are kept for the next iteration of the algorithm without removing, that in turn, effects the stability. Although works in literature report an increase in the accuracy of this classification, the results in Figure~\ref{fig:stability} shows that the stability suffers.

We can think of different alternatives to compensate this difficulty by tweaking the Ant-Miner classification procedure. But, an acclaimed solution that dealt with the unstable classifiers in the traditional machine learning is the use of ensembles. Ensemble of classifiers improves the stability as well as the accuracy. Before we delve into the proposed ensemble approach we take a quick walk through the state-of-the-art ACO based ensemble classifiers.

\subsection{ACO ensembles}
\label{ensembles}
We found a few number of ensemble ACO classifiers in literature. Of which, Chen $\&$ Wong~\cite{Chen:stacking} presented an ACO approach for the stacking ensemble, named ACO-Stacking. It addressed the difficulty to determine the configuration of base-layer classifiers and the meta-layer classifiers in Stacking. They demonstrated improved classification accuracy by selecting optimal configuration of base-classifiers using ACO search process. 

Khalid $\&$ Freitas~\cite{Khalid:ensemble} proposed a new ACO based Bayesian network ensembles with a new ABC-Miner~\cite{Khalid:ABCMiner} as the base classifier, in order to predict the ageing related protein functions in bioinformatics. They identified the importance of human ageing related protein function prediction as a major application of hierarchical classification. They tackled this prediction task with the construction of a meta classifier through the best classifier selection, majority and weighted voting.

\begin{algorithm}
	\caption{Pseudo code for the ensemble Ant-Miner algorithm.}
	\begin{algorithmic}[1]
		\REQUIRE \textit{S}, Original training examples; \\
		\textit{I}, Base Ant-Miner; \\
		\textit{T}, Number of learning iterations; \\
		\ENSURE bagged model
		\FORALL{$t$ such that $1\leq t \leq \textit{T}$}
		\STATE $S_{t}$ $\leftarrow$ \textit{Bootstrap(S)}; \COMMENT{Generate a bootstrap sample with replacement from \textit{S}}
		\STATE $h_{t}$ $\leftarrow$ \textit{L}($S_{t}$); \COMMENT{Train base Ant-Miner on sample data $S_{t}$}
		\STATE h[t] $\leftarrow$ $h_{t}$; \COMMENT{Store the ant discovered model}
		\ENDFOR
		\WHILE{instance $\leq$ maximum examples}
		\FORALL{$t$ such that $1\leq t \leq sizeOf(h)$} 
		\STATE prediction[t] $\leftarrow$ classifyInstance(instance,h[t]); \COMMENT{Classify the instance across all the generated models.}
		\ENDFOR
		\RETURN majorityVoting(prediction); \COMMENT{return the class that's predicted most.}
		\ENDWHILE
	\end{algorithmic} \label{alg:ensemble}
\end{algorithm}

Despite the proposed approach in this paper, to the best of our knowledge there are few investigations in improving the stability of Ant-Miners. We found that there is a lot of scope to produce ensemble Ant-Miners to reduce the classification error there by improving the stability that ultimately helps to improve the classification ability of Ant-Miners.

\section{eAnt-Miner}
\label{eantminer}
The proposed ensemble Ant-Miner algorithm follows the traditional bagging procedure explained in~\cite{Breiman:bagging}. Bagging is a statistical technique defined as the bootstrapped aggregation where the training set is bootstrapped for \textit{n} times at which, \textit{n} classification models are constructed. Then, each instance in the data set is classified across all the models as a result, the predictions are aggregated with the majority voting in case of classification, averaged if it is applied on regression. Algorithm~\ref{alg:ensemble} presents the pseudo code of the ensemble Ant-Miner. 

In summary, the eAnt-Miner works as follows. It starts by taking the data set (\textit{S}), base classifier (\textit{I}), number of learning iterations (\textit{T}) as the input. At any given instance the base classifier will be one variant of the state-of-the-art Ant-Miner classification algorithms considered in this paper. This algorithm works in two major steps. In the first step, it constructs a bootstrap sample (\textit{$S_t$}) with a random replacement from the training data. In the second step, a model is prepared on the bootstrap sample (\textit{$S_{t}$}), then, the newly generated model is stored for testing. This procedure is repeated until it reached the maximum number of learning iterations (\textit{T}) so that \textit{T} Ant-Miner classification models are resulted. Now, each instance is classified on all the \textit{T} models that record the predicted class. The predicted class of the instance is decided with a majority voting on all the predictions produced by \textit{T}  different models. For example: in case of binary classification, if we consider $5$  predictions [\textit{yes,no,yes,no,yes}] then, \textit{yes} is the predicted class resulted from the majority voting as it counts to $3$.

Note that the original bagging procedure reduces the variance of the classifier as a result the predictive power of the classification improves. We discuss the impact of the bagged ensemble in the following sections.

  \begin{table} [htp]
    \caption{The four experimented Ant-Miner variations and their experimental parameter settings.}
    \begin{center}
      \begin{tabular}{l c c c l c l}
        \specialrule{.1em}{.03em}{.03em}
        Abbreviation & & & &  Classifier & Ref.  &  Parameter Description \\
        \specialrule{.1em}{.03em}{.03em}
        cAM & & & &  cAntMiner & \cite{fernando:amcontinous} & The number of ants are set to $3000$ while \\ 
        & & & & & & the minimum number of data points per rule \\
        & & & & & & are $5$ whereas the maximum uncovered data \\
        & & & & & & points are $10$ and convergence limit on the \\
        & & & & & & number of rules is $10$.  \\
        $cAM_{PB}$ & & & & cAntMiner$_{PB}$ & \cite{Medland:campb} & The number of ants in a colony is set to $5$ \\ 
        & & & & & & with a maximum number of iterations of $500$, \\
        & & & & & & an evaporation factor of $0.90$, the minimum \\ 
        & & & & & & covered data points are $10$. An error based \\
        & & & & & & rule list quality function has been used.  \\
        ATM  & & & & Ant-Tree-Miner & \cite{fernando:trees} & The colony size is set to $50$, the maximum \\
        & & & & & & number of iterations is $500$, the evaporation \\
         & & & & & & factor is $0.90$ with the minimum number \\
         & & & & & & of iterations per branch is $3$. An error based \\
         & & & & & & tree measure is used to determine the quality \\ 
        & & & & & & of the ant discovered tree.   \\
        U-cAM$_{PB}$ & & & & Unordered-cAntMiner$_{PB}$ & \cite{fernando:ucampb} & The colony size of $5$, a maximum number \\
        & & & & & & of iterations is set to $500$, evaporation factor \\ 
        & & & & & & is $0.90$. The dynamic rule quality function is \\ 
         & & & & & & the  best method to experiment hence, it is \\
        & & & & & & used in this paper.\\
       \specialrule{.1em}{.03em}{.03em}
      \end{tabular} \label{tab:params}
    \end{center}
  \end{table}  

\section{Experiments}
\label{experiments}
We consider four different Ant-Miner classifiers that are discussed in section~\ref{stability}. The same classifiers are used as the base classifiers with the proposed ensemble approach in which the names are preceded with the letter ($e$) from here onwards. Table~\ref{tab:params} presents all the four selected Ant-Miner variations. Column $1$ is the abbreviation of the corresponding Ant-Miner which we use to refer to that particular algorithm from here till the end of the paper. It also provides a brief explanation about the parameter settings, all the algorithms used the default parameter values proposed by the corresponding authors. A detailed explanation of the parameters can be found in their respective citations.

  \begin{table} [htp]
    \caption{Summary of datasets used in the experiments.}
    \begin{center}
      \begin{tabular}{l c c c c c c c l r r r r r}
        \specialrule{.1em}{.03em}{.03em}
        \multirow{2}{8pt}{Abbreviation} & & & & & & & & Dataset  & \multicolumn{3}{c}{\# Attributes} & \hspace{0.5 cm} \multirow{2}{35pt}{\# Classes}  & \multirow{2}{45pt}{\# Examples} \\
        \cline{10-12}
        	& & & & & & & & Description & Nominal & & Continuous &  	&  \\
        \specialrule{.1em}{.03em}{.03em}
        anneal & & & & & & & & annealing & 29 & & 9 &  6 & 898  \\
        balance & & & & & & & & balance scale & 4 & & 0 &  3 & 625 \\
        breast-l & & & & & & & & breast cancer ljubljana & 9 & & 0  &  2 & 286 \\
        breast-t & & & & & & & & breast tissue & 0 & & 9 & 6 & 106 \\
        breast-w & & & & & & & & breast cancer wisconsin & 0 & & 30 & 2 & 569 \\
        credit-a & & & & & & & & credit approval & 8 & & 6 & 2 & 690 \\
        cylinder & & & & & & & & cylinder bands & 16 & & 19 & 2 & 540 \\
        derm & & & & & & & & dermatology & 33 & & 1 & 6 & 366 \\
        glass & & & & & & & & glass & 0 & & 9 & 7 & 214 \\
        heart-c & & & & & & & & heart cleveland & 6 & & 7 & 5 & 303 \\
        heart-h & & & & & & & & heart hungarian & 6 & & 7 & 5 & 294 \\
        ionos & & & & & & & & ionosphere & 0 & & 34 & 2 & 351 \\
        iris & & & & & & & & iris & 0 & & 4 & 3 & 150 \\
        liver & & & & & & & & liver disorders & 0 & & 6 & 2 & 345 \\
        park & & & & & & & & parkinsons & 0 & & 22 & 2 & 195 \\
        pima & & & & & & & & pima & 0 & & 8 & 2 & 768 \\
        wine & & & & & & & & wine & 0 & & 13 & 3 & 178 \\
        \specialrule{.1em}{.03em}{.03em}
      \end{tabular} \label{tab:dataset}
    \end{center}
  \end{table} 
  
We performed two different type of experiments: the first one deals with the single classifier classification, the next one contains the ensemble of the classifiers. In the case of single Ant-Miner classification experiments, the performance is measured with the help of $10$-fold cross-validation where the given data set is divided into approximately ten equal stratified partitions. One partition is held out as test data while the classifier learns to discover a model on the remaining nine partitions. In this case, the miss-classification rate (i.e., the sum of the false positives and false negatives) is measured on the unseen (hold-out) partition considering that as the test error. The final value of the test error for a particular dataset is the average value obtained across the ten folds.

On the other hand, in the case of ensemble classification, we split the input data set into both training ($70\%$) and test ($30\%$) sets. Bootstrapped random samples with replacement are created on the training set. i.e., We created $10$ random samples of data with the size of the training set. As a result, the base Ant-Miner discovered ten different models that are used to created a single bagged ensemble classifier. The class values of the test set (unseen data) are predicted using the majority voting procedure (as explained in section~\ref{ensembles}). In this case also we recorded the test error of the resultant bagged ensemble. Note that the test error is often used as the classification error that explains the generalization ability of the classifier. This process is repeated for $10$ iterations so that the final value is the average of all the iterations.

  \begin{table} [htp]
      \caption{Average test error (in $\%$) measured by applying 10-fold cross-validation in the case of single classifiers and $10$ random bootstrapped samples with replacement for $10$ iterations in the case of ensemble classification. The algorithm with the lowest error rate is shown in bold.}    
      \begin{center}
        \begin{tabular}{l c c c c c c c c c c c c c c c c}
          \specialrule{.1em}{.03em}{.03em}
           Dataset & & cAM & & ecAM & & $cAM_{PB}$ & & $ecAM_{PB}$  & & ATM & & eATM & & U-cAM$_{PB}$ & & eU-cAM$_{PB}$ \\
          \specialrule{.1em}{.03em}{.03em}
          anneal & & 10.37 & & 9.92 & & 6.43 & & 5.69 & & 1.99 & & \textbf{1.67} & & 4.32 & & 2.18 \\
          balance & & 31.86 & & 30.14 & & 21.59 & & \textbf{18.24} & & 36.79 & & 32.16 & & 23.59 & & 21.06 \\
          breast-l & & 29.07 & & 27.93 & & 28.26 & & 25.41 & & 25.80 & & \textbf{20.40} & & 27.24 & & 21.85 \\
          breast-t & & 34.18 & & 31.25 & & 30.28 & & 29.63 & & 32.27 & & \textbf{28.15} & & 36.09 & & 26.88 \\
          breast-w & & 12.25 & & 11.64 & & 8.73 & & \textbf{4.52} & & 9.86 & & 6.14 & & 9.43 & & 7.11 \\
          credit-a & & 18.21 & & 18.57 & & 16.80 & & 14.19 & & 13.48 & & \textbf{11.57} & & 17.29 & & 15.44 \\
          cylinder & & 34.16 & & 30.79 & & 29.67 & & 27.19 & & 25.64 & & \textbf{24.13} & & 29.15 & & 26.23 \\
          derm & & 14.42 & & 11.30 & & 10.27 & & 8.60 & & 7.39 & & \textbf{4.56} & & 12.16 & & 11.89 \\
          glass & & 40.46 & & 38.04 & & 34.18 & & \textbf{30.93} & & 35.59 & & 36.22 & & 36.10 & & 33.48 \\
          heart-c & & 47.16 & & 43.97 & & 35.49 & & \textbf{32.15} & & 39.94 & & 36.16 & & 38.42 & & 36.69 \\
          heart-h & & 38.29 & & 36.42 & & 37.18 & & 35.56 & & 32.31 & & \textbf{29.87} & & 36.46 & & 34.19 \\          
          ionos & & 18.87 & & 16.12 & & 11.05 & & 10.46 & & 9.82 & & \textbf{7.51} & & 10.59 & & 9.56 \\
          iris & & 10.67 & & 5.56 & & 8.87 & & 7.41 & & 5.38 & & \textbf{3.70} & & 7.46 & & 6.45 \\
          liver & & 38.54 & & 35.12 & & 36.65 & & 32.51  & & 30.28 & & 29.18 & & 33.01 & & \textbf{26.81} \\
          park & & 19.95 & & 16.25 & & 14.59 & & 11.14 & & 11.18 & & \textbf{7.60} & & 14.31 & & 12.82 \\
          pima & & 28.61 & & 29.73 & & 26.48 & & \textbf{24.17} & & 27.42 & & 26.27 & & 25.96 & & 24.87 \\
          wine & & 11.76 & & 9.26 & & 7.41 & & 3.69 & & 4.11 & & 3.85 & & 5.62 & & \textbf{3.73} \\
          \specialrule{.1em}{.03em}{.03em} 
        \end{tabular} \label{tab:error}
      \end{center}
  \end{table}

    \begin{table} [htp]
      \caption{Average number of terms (average [standard error]) discovered by the standard Ant-Miner and ensemble Ant-Miners in classifying a data point. The lowest value is in boldface for a given dataset.}    
      \begin{center}
        \begin{tabular}{l c c c c c c c c c c c c c}
          \specialrule{.1em}{.03em}{.03em}
           Dataset & & cAM & & ecAM & & $cAM_{PB}$ & & $ecAM_{PB}$ & & U-cAM$_{PB}$ & & eU-cAM$_{PB}$ \\
          \specialrule{.1em}{.03em}{.03em}	
			anneal & & 13.36 [0.45] & & 15.78 [2.14] & & 7.84 [0.39] & & 9.15 [1.11] & & \textbf{2.34 [0.03]} & & 4.16 [1.05] \\
			balance & & 11.66 [0.05] & & 14.17 [2.31] & & 6.51 [0.26] & & 9.46 [0.89] & & \textbf{3.4 [0.01]} & & 5.17 [1.06] \\
			breast-l & & 9.13 [0.23] & & 11.46 [2.56] & & 6.28 [0.13] & & 7.23 [1.46] & & \textbf{2.44 [0.08]} & & 4.16 [1.29] \\
			breast-t & & 6.14 [0.04] & & 7.73 [2.18] & & 3.93 [0.01] & & 4.29 [1.08] & & \textbf{1.91 [0.11}] & & 2.15 [1.18] \\
			breast-w & & 12.15 [0.56] & & 14.46 [1.57] & & 9.31	[0.10] & & 10.42 [1.38] & & \textbf{3.41 [0.07]} & & 4.56	[0.53] \\
			credit-a & & 16.34 [0.67] & & 17.06 [1.03] & & 10.26 [0.01] & & 11.34 [1.07] & & \textbf{4.56 [0.06]} & & 6.11 [0.29] \\
			cylinder & & 29.17 [1.26] & & 31.38 [2.16] & & 15.29 [0.15] & & 17.16 [1.10] & & \textbf{3.48 [0.03]} & & 5.16 [0.19] \\
			derm & & 17.48 [0.21] & & 18.49 [1.36] & & 16.26 [0.21] & & 17.09 [1.67] & & \textbf{5.46 [0.09]} & & 6.18 [0.13] \\
			glass & & 8.97 [0.01] & & 10.19 [0.46] & & 6.79 [0.71] & & 7.45	[0.18] & & \textbf{2.71 [0.11] }& & 3.38 [0.26] \\
			heart-c & & 16.32 [1.05] & & 16.84 [1.16] & & 10.49 [0.49] & & 11.14 [0.56] & & \textbf{3.46 [0.01]} & & 4.16 [0.12] \\
			heart-h & & 9.81 [0.84] & & 9.16 [1.18] & & 6.13 [0.16] & & 7.43 [0.56] & & \textbf{2.98 [0.19]} & & 4.19 [0.09] \\
			ionos & & 9.12 [0.84] & & 11.14 [1.09] & & 8.15	[0.16] & & 9.28	[1.06] & & \textbf{3.24 [0.09]} & & 5.02 [0.06] \\
			iris & & 2.74 [0.02] & & 5.16 [0.16] & & 1.26 [0.04] & & 3.19 [0.42] & & \textbf{0.86 [0.06]} & & 1.96 [0.10] \\
			liver & & 15.61	[0.05] & & 18.82 [1.35] & & 10.9 [0.01] & & 12.18 [0.46] & & \textbf{2.31 [0.24]} & & 4.56 [1.59] \\
			park & & 1.4 [0.07] & & 3.26 [0.24] & & \textbf{0.96 [0.04]} & & 1.63 [0.24] & & 1.8	[0.10] & & 3.05	[0.37] \\
			pima & & 10.08	[1.52] & & 12.16 [2.64] & & 6.34 [0.11] & & 8.96 [1.19] & & \textbf{2.01 [0.05]} & & 4.56 [0.18] \\
			wine & & 5.59 [0.07] & & 7.28 [2.43] & & 3.9 [0.34] & & 5.63 [1.19] & & \textbf{1.26 [0.10]} & & 3.17	[0.19] \\
          \specialrule{.1em}{.03em}{.03em}
        \end{tabular} \label{tab:termsize}
      \end{center}
    \end{table} 

\subsection{Results}
In order to perform the experiments, we have designated $17$ publicly available datasets from the UCI Machine Learning repository~\cite{Bache:uci}. Table~\ref{tab:dataset} expounds the data sets used in the experiments. All the data sets contain unique data points (no duplicates). We compare the experimental results obtained in the two set-ups	 (single, ensemble classification) that determine the effect of the proposed approach on the classification error. 

Table~\ref{tab:error} represents the comparative analysis of the average test error rates of single Ant-Miners with that of the ensemble Ant-Miners (bagged Ant-Miners). The lower the error value the better the classifier performance. A value for a given data set is shown in bold if it is significantly lower than the rest of the algorithms. Since, the Ant-Miners are stochastic algorithms we executed single Ant-Miner classifiers for $10$ iterations with each iteration having one $10$-fold cross-validation (i.e., 10 x 10 times for each data set). The average error value is computed across all the ten iterations. In case of ensemble classification, the average test error is calculated for $10$ iterations with each iteration containing $10$ random bootstrapped samples (i.e., 10 x 10 times for each data set). 

\begin{table}[htp]
	\centering
	\caption{Statistical test results of the algorithms on average test error as per the non-parametric Friedman test with Hommel's post hoc test. Significantly different results at $\alpha$ = $0.05$ are in boldface.}
	\begin{tabular}{cccccccccccccccc}
		\specialrule{.1em}{.03em}{.03em}
		Algorithm & & & & & Avg. Rank & & & & & $p$-value & & & & & $p$-Hommel\\
		\specialrule{.1em}{.03em}{.03em}
		$eUcAM_{pb}$ & & & & & \textbf{6.511302390730246} & & & & & \textbf{7.450199199902389E-11} & & & & & \textbf{0.0071428571428571435}\\
		$ecAM_{pb}$ & & & & & \textbf{5.951190357119043} & & & & & \textbf{2.661992555310251E-9} & & & & & \textbf{0.008333333333333333}\\
		eATM & & & & & \textbf{5.811162348716241} & & & & & \textbf{6.204056085147265E-9} & & & & & \textbf{0.01}\\
		$UcAM_{pb}$ & & & & & \textbf{4.62092427729243} & & & & & \textbf{3.820342059789907E-6} & & & & & \textbf{0.0125}\\
		$cAM_{pb}$ & & & & & \textbf{3.010602180660221} & & & & & \textbf{0.0026073020496656255} & & & & & \textbf{0.016666666666666666}\\
		ATM & & & & & \textbf{2.940588176458821} & & & & & \textbf{0.0032758974829086296} & & & & & \textbf{0.025}\\
		ecAM & & & & & 1.6803361008336113 & & & & & 0.09289194088370535 & & & & & 0.05\\
		\specialrule{.1em}{.03em}{.03em}
	\end{tabular}\label{tab:significance}
\end{table}

Table~\ref{tab:termsize} summarizes the results obtained in terms of the average size of the discovered model. Each value in the table represents the average number of terms in the discovered model followed by the standard deviation (average [standard deviation]) for the corresponding data set. Smaller the average simpler the model, significantly small models are shown in boldface. In order to prove the significance of the reported results, we performed statistical significance tests at $\alpha$ = $0.05$. Since the bootstrapped sample of data sets contribute to discover a model hence, the size of the discovered model is increased with the new ensemble approach.

Table~\ref{tab:significance} presents the results based on Friedman tests with Hommel's post-hoc analysis~\cite{Demsar:scc,Garcia:sccex} for the test error and the model size. For each algorithm, the table shows the average rank, \textit{p}-value, and the adjusted $p_{Homm}$ value. The statistically significant difference is determined with the adjusted p-value: if the $p_{Hommel}$ is less than $0.05$ then the difference in the ranks is statistically significant at $\alpha$ = $0.05$ level. All the ensembles except $ecAM$, achieved better ranks in comparison with the single classifiers improved the performance by reducing the test error. Of those three ensembles, $eU-cAM_{PB}$ attained the best average rank outperforming the state-of-the-art Ant-Miners, thus, improved the performance. $ecAM$ is the only bagged ensemble that has not improved the performance over the standard Ant-Miner variations.

\begin{table}[htp]
	\centering
	\caption{Shaffer's multiple hypothesis comparison test results at $\alpha=0.05$}
	\begin{tabular}{cccc}
		\specialrule{.1em}{.03em}{.03em}
		Algorithms & Avg. Rank & $p$-value  & $p$-Shaffer\\
		\specialrule{.1em}{.03em}{.03em}
		cAM vs. $eUcAM_{pb}$ & \textbf{6.5113023907} & \textbf{7.450199199E-11} & \textbf{0.00178571428}\\
		cAM vs. $ecAM_{pb}$ & \textbf{5.9511903571} & \textbf{2.66199255531E-9}  & \textbf{0.00238095238}\\
		cAM vs. eATM & \textbf{5.8111623487} & \textbf{6.204056085E-9}  & \textbf{0.00238095238}\\
		ecAM vs. $eUcAM_{pb}$ & \textbf{4.83096628989} & \textbf{1.35872009979E-6}  & \textbf{0.0023809523}\\
		cAM vs. $UcAM_{pb}$ & \textbf{4.6209242772} & \textbf{3.82034205978E-6}  & \textbf{0.0023809523}\\
		ecAM vs. $ecAM_{pb}$ & \textbf{4.2708542562} & \textbf{1.947256374517E-5}  & \textbf{0.002380952}\\
		ecAM vs. eATM & \textbf{4.1308262478} & \textbf{3.61461714458E-5}  & \textbf{0.00238095238}\\
		$eUcAM_{pb}$ vs. ATM & \textbf{3.57071421427} & \textbf{3.56009163215E-4}  & \textbf{0.00238095238}\\
		$cAM_{pb}$ vs. $eUcAM_{pb}$ & \textbf{3.50070021007} & \textbf{4.640375320753E-4}  & \textbf{0.003125}\\
		$ecAM_{pb}$ vs. ATM & \textbf{3.0106021806} & \textbf{0.00260730204966}  & \textbf{0.003125}\\
		cAM vs. $cAM_{pb}$ & \textbf{3.01060218066} & \textbf{0.00260730204966}  & \textbf{0.003125}\\
		$cAM_{pb}$ vs. $ecAM_{pb}$ & \textbf{2.940588176458} & \textbf{0.003275897482}  & \textbf{0.003125}\\
		cAM vs. ATM & \textbf{2.94058817645} & \textbf{0.00327589748290}  & \textbf{0.00312}\\
		ecAM vs. $UcAM_{pb}$ & \textbf{2.94058817645} & \textbf{0.003275897482}  & \textbf{0.003333333333}\\
		ATM vs. eATM & \textbf{2.87057417225} & \textbf{0.0040972705955}  & \textbf{0.003571428571}\\
		$cAM_{pb}$ vs. eATM & \textbf{2.8005601680} & \textbf{0.005101399646}  & \textbf{0.0038461538461}\\
		$UcAM_{pb}$ vs. $eUcAM_{pb}$ & 1.89037811343 & 0.058707408431 & 0.00416666666\\
		cAM vs. ecAM & 1.680336100833 & 0.092891940883  & 0.00454545454\\
		$UcAM_{pb}$ vs. ATM & 1.6803361008 & 0.092891940883  & 0.005\\
		$cAM_{pb}$ vs. $UcAM_{pb}$ & 1.610322096632 & 0.107327557761  & 0.0055555555555\\
		$ecAM_{pb}$ vs. $UcAM_{pb}$ & 1.33026607982 & 0.183430618982  & 0.00625\\
		ecAM vs. $cAM_{pb}$ & 1.330266079826 & 0.18343061898  & 0.0071428571428\\
		ecAM vs. ATM & 1.2602520756252091 & 0.2075784423  & 0.008333333333\\
		$UcAM_{pb}$ vs. eATM & 1.190238071423 & 0.233952833  & 0.01\\
		$eUcAM_{pb}$ vs. eATM & 0.700140042014 & 0.4838398  & 0.0125\\
		$ecAM_{pb}$ vs. $eUcAM_{pb}$ & 0.56011203 & 0.575403022  & 0.016666666\\
		$ecAM_{pb}$ vs. eATM & 0.14002800840280155 & 0.8886378608  & 0.025\\
		$cAM_{pb}$ vs. ATM & 0.07001400420140025 & 0.9441825132  & 0.05\\
		\specialrule{.1em}{.03em}{.03em}
	\end{tabular} \label{tab:shaffer}
\end{table}

Shaffer~\cite{Shaffer:multiple} presented a statistical procedure in order to perform pair-wise comparison of multiple hypothesis. In this paper, we use this procedure to conduct a pair-wise comparison of multiple Ant-Miners. Table~\ref{tab:shaffer} presents the multi-classifier comparison results at $\alpha$=0.05. In the table, significantly better results are shown in boldface. For example, in \textit{cAM vs. $eUcAM_{pb}$}, from the Shaffer's test, \textit{$eUcAM_{pb}$} outperforms \textit{cAM}. This analogy applies to all the pair-wise comparisons that their $p$-Shaffer value is in bold.

Recall that the Ant-Tree-Miner and the Unordered-cAnt-Miner$_{PB}$ are the most unstable (discussed in section~\ref{stability}). Finally, from the results, it is now clear that the performance of the Ant-Miners is improved, thus, the stability also enhanced, through the introduction of the ensemble approach. 

\section{Conclusion}
\label{conclusion}
This paper has proposed to apply the bagged ensemble approach on Ant-Miners, eAnt-Miner, to improve the stability of ACO classification. The stochastic nature of the ACO classifiers exert instability in the final discovered model that contributes to the high level of classification error. This is an important reason for the instability of the Ant-Miners. In the case of Ant-Tree-Miner, change in at least one node of a tree highly changes the final results, which is another important reason for instability. The main objective is to reduce the error there by improving the accuracy of various Ant-Miners that indeed improves the stability.

We compared the proposed ensemble approach with the single Ant-Miners in terms of average test error as it is often used as a standard measure to evaluate the performance of a classifier. We evaluated the new approach on $17$ publicly available data sets. As a result, we reduced the test error of the classifiers that improved the stability and the performance of the ACO classification. Our results showed that the ensemble Ant-Tree-Miner (eATM) outperformed the state-of-the-art Ant-Miners and most of the proposed ensemble approaches. 

In general, ensemble methodology seeks opinion of several experts before taking a decision on the prediction. Researchers of various fields starting from statistics, evolutionary computation, to machine learning have explored the use of ensemble methods. The proposed approach concentrated only on producing bagged ensembles for multiple and binary class classification. There are other ensemble approaches $BOOST$, $ADABOOST$ that we aim to explore in future. We recommend to apply these new methods to hierarchical data sets. We also recommend to employ the ensemble methods proposed in~\cite{Khalid:ensemble,Chen:stacking} on Ant-Miner variations to further improve the ACO based classification.

\bibliographystyle{abbrv}
\bibliography{references}

\end{document}